\documentclass[conference]{IEEEtran}
\IEEEoverridecommandlockouts
\usepackage{xcolor}
\usepackage{amsmath,amssymb,amsfonts}
\usepackage{graphicx}
\usepackage{textcomp}

\begin{document}
\title{Analytical Logit Scaling for High-Resolution Sea Ice Topology Retrieval from Weakly Labeled SAR Imagery}

\author{
\IEEEauthorblockN{Reda Elwaradi, Julien Gimenez, Stéphane Hordoir, and Mehdi Ait Hamma}
\IEEEauthorblockA{\textit{Mastère Spécialisé IA} \\
\textit{Télécom Paris, Institut Polytechnique de Paris}\\
Palaiseau, France \\
reda.elwaradi@telecom-paris.fr}
\\
\IEEEauthorblockN{Adrien Chan Hon Tong\textsuperscript{*} and Flora Weissgerber\textsuperscript{*}}
\IEEEauthorblockA{\textit{DTIS} \\
\textit{ONERA, Université Paris-Saclay}\\
91120, Palaiseau, France \\
*Supervision only}
}

\maketitle

\begin{abstract}
High-resolution sea ice mapping using Synthetic Aperture Radar (SAR) is crucial for Arctic navigation and climate monitoring. However, operational ice charts provide only coarse, region-level polygons (weak labels), forcing automated segmentation models to struggle with pixel-level accuracy and often yielding under-confident, blurred concentration maps. In this paper, we propose a weakly supervised deep learning pipeline that fuses Sentinel-1 SAR and AMSR-2 radiometry data using a U-Net architecture trained with a region-based loss. To overcome the severe under-confidence caused by weak labels, we introduce an Analytical Logit Scaling method applied post-inference. By dynamically calculating the temperature and bias based on the latent space percentiles (2\% and 98\%) of each scene, we force a physical binarization of the predictions. This adaptive scaling acts as a topological extractor, successfully revealing fine-grained sea ice fractures (leads) at a 40-meter resolution without requiring any manual pixel-level annotations. Our approach not only resolves local topology but also perfectly preserves regional macroscopic concentrations, achieving a 78\% accuracy on highly fragmented summer scenes, thereby bridging the gap between weakly supervised learning and high-resolution physical segmentation.
\end{abstract}


\section{Introduction}
Sea ice is a critical indicator of climate change and a major factor in Arctic maritime navigation. High-resolution Synthetic Aperture Radar (SAR), such as the Sentinel-1 mission, provides all-weather, day-and-night imaging capabilities, making it the primary data source for operational sea ice charting. Recently, deep learning architectures, particularly Convolutional Neural Networks (CNNs), have demonstrated immense potential in automating sea ice parameter retrieval, heavily spurred by initiatives like the AI4Arctic and AutoICE challenges \cite{chen2024weakly, stokholm2022ai4arctic}.

However, training these models requires large volumes of annotated data. Operational ice charts, produced by human analysts based on the World Meteorological Organization (WMO) Egg Code \cite{wmo2014sea}, represent sea ice concentration (SIC) as macroscopic averages over large polygons. This creates a severe scale mismatch known as the ``average pixel illusion'': a polygon labeled with 70\% SIC indicates a regional proportion, not that every single pixel inside is composed of 70\% ice. Consequently, treating these polygons as pixel-level ground truth provides only ``weak labels'' for training high-resolution segmentation models.

Recent studies have attempted to address this challenge. For instance, Chen et al. \cite{chen2024weakly} successfully implemented a region-based loss function to train a U-Net model under weak supervision. While this approach effectively estimates macroscopic concentration, the resulting maps often suffer from systemic under-confidence, producing blurred transitions that fail to resolve fine-grained topologies. Conversely, Dai et al. \cite{dai2026panarctic} demonstrated that CNNs can achieve exceptional topological precision---specifically in detecting narrow sea ice fractures known as \textit{leads}. However, their approach necessitates the meticulous, manual annotation of these leads at the pixel level, an expensive and subjective process that is difficult to scale pan-Arctic.

In this paper, we propose a novel approach that bridges the gap between the scalability of weakly supervised learning and the topological precision of fully supervised methods. By acknowledging the latent under-confidence of networks trained on regional weak labels, we introduce an Analytical Logit Scaling technique applied post-inference. Instead of learning scaling parameters via gradient descent---which is hindered by weak labels---our method dynamically computes the optimal Temperature and Bias based on the statistical percentiles of each scene's latent space (logits). This adaptive scaling explicitly forces a physical binarization of the radar backscatter, revealing high-resolution topological features such as leads without requiring any manual pixel-level annotations, while strictly preserving the macroscopic concentration of the region.

\section{Data and Problem Formulation}

\subsection{AI4Arctic Dataset and Multimodal Fusion}
The dataset used in this study is derived from the AI4Arctic Sea Ice Challenge. The input pipeline fuses high-resolution Sentinel-1 Synthetic Aperture Radar (SAR) imagery with low-resolution AMSR-2 passive microwave radiometry. The input tensor consists of four channels designed to maximize spectral and contextual information. The SAR channels (HH and HV polarization) at a nominal 40-meter resolution capture surface roughness and volume scattering, effectively distinguishing deformed ice from calm water. The AMSR-2 brightness temperature channel resolves physical ambiguities, particularly in ocean areas where wind-induced waves mimic the SAR backscatter of ice. 

To handle the high dynamic range of these sensors, we apply a robust scaling strategy: SAR backscatter is clipped to $[-30, 20]$ dB, and AMSR-2 temperature to $[150, 300]$ K, before standard normalization. The ground truth consists of operational Ice Charts, rasterized into Sea Ice Concentration (SIC) maps ranging from $0$ (pure open water) to $1.0$ (solid ice).

\subsection{The Spatial Paradox of Weak Labels}
In operational Ice Charts, SIC is annotated by human experts as macroscopic averages over large geographical polygons based on the WMO Egg Code. This creates a fundamental scale mismatch for pixel-level segmentation. Let $P$ be a macroscopic polygon with an assigned concentration $C_p \in [0, 1]$. In standard learning, a network predicting a value $\hat{y}_i$ for a pixel $i \in P$ is penalized by comparing it directly to $C_p$. 

However, at a 40-meter resolution, a pixel physically contains either solid ice ($y_{\text{true}} = 1$) or open water ($y_{\text{true}} = 0$). By forcing the network to predict $\hat{y}_i = C_p$ (e.g., $0.7$), the training loss traps the model in a state of severe local under-confidence. The network learns to predict intermediate probabilities everywhere to safely minimize the regional error. While this preserves the global average, it results in a blurred probability map that completely destroys high-frequency spatial topologies, making it impossible to map narrow sea ice fractures (leads).

\section{Methodology}

\subsection{Base Architecture and Region-based Loss}
The foundation of our pipeline is a U-Net architecture \cite{ronneberger2015u} adapted for multimodal regression. The encoder extracts multi-scale spatial features through successive \texttt{DoubleConv} blocks, while the decoder reconstructs the spatial resolution via skip connections. Instead of a final sigmoid activation directly penalizing pixel-wise predictions, the network outputs a dense probability map which is aggregated via a spatial Global Average Pooling (GAP) \cite{lin2013network} over the input patch. The network is trained using a weakly supervised loss function that minimizes the difference between the patch's area-weighted macroscopic concentration $C_p$ and this pooled prediction. Crucially, this is supplemented by a soft binarization regularization term, which forces the network to break spatial symmetry and build topological gradients in the latent logit space $Z$ without violating the regional constraint.

\subsection{Spatial Regularization}
SAR imagery is inherently corrupted by speckle noise, which introduces high-frequency artifacts in the latent space. Before any calibration, we extract the raw logit map $Z$ generated by the trained U-Net during inference and apply a spatial low-pass filter (Gaussian Blur). This spatial regularization smooths out the speckle-induced noise while preserving the strong structural gradients corresponding to actual sea ice features and leads.

\subsection{Analytical Logit Scaling}
The core contribution of this work lies in overcoming the under-confidence induced by the region-based loss. Standard Platt Scaling or Temperature Scaling \cite{guo2017calibration} adjusts the logits $z$ using a Temperature $T$ and a bias $b$ prior to the sigmoid activation $\sigma$:

\begin{equation}
\hat{p} = \sigma \left( \frac{z - b}{T} \right)
\label{eq:platt}
\end{equation}

Typically, $T$ and $b$ are optimized via gradient descent on a perfectly labeled, hold-out validation set. However, in our weakly supervised operational paradigm, pixel-level ground truth is entirely unavailable. Attempting to optimize these scaling parameters via gradient descent using the same macroscopic weak labels would merely force the model back into its under-confident state to minimize the regional error.

To bypass this, we propose an Analytical Logit Scaling method that computes $T$ and $b$ dynamically for each inference scene based on the statistical distribution of its spatially smoothed logits. To ensure robustness against outliers, we extract the 2nd and 98th percentiles of the logit map, denoted respectively as $z_{2\%}$ and $z_{98\%}$. 

Our objective is to explicitly force a physical binarization by stretching this latent dynamic range to the interval $[-5, 5]$, which corresponds to the saturation bounds of the sigmoid function ($\sigma(-5) \approx 0.006$ and $\sigma(5) \approx 0.993$). The analytical parameters are thus defined as:

\begin{equation}
b = \frac{z_{98\%} + z_{2\%}}{2}
\label{eq:bias}
\end{equation}

\begin{equation}
T = \frac{z_{98\%} - z_{2\%}}{10}
\label{eq:temp}
\end{equation}

By applying these dynamic parameters, the logits are re-centered and stretched. The subsequent sigmoid activation acts as a hard thresholding function, effectively converting the blurred concentration map into a high-resolution, binary topological segmentation of solid ice and open water, directly mitigating the weak label effect.

\subsection{Operational Large-Scale Inference}
Applying a CNN to full-swath Sentinel-1 scenes (typically spanning hundreds of kilometers) introduces spatial discontinuities if processed through naive non-overlapping tiling. To ensure seamless topological continuity, we implemented a sliding window inference strategy. The $256 \times 256$ inference window moves with a stride of 64 pixels. Consequently, each pixel in the core scene is inferred 16 times under different spatial contexts. The final logit map $Z$ is computed as the arithmetic mean of these overlapping predictions.

Despite the heavy overlapping (157 large patches per scene), the approach is computationally highly efficient. The Analytical Logit Scaling adds virtually zero computational overhead, as the dynamic percentile extraction and linear stretching are vectorized $O(N)$ operations. The entire pipeline processes a full high-resolution SAR scene in approximately 22 minutes on a consumer-grade Apple Silicon processor (via the MPS backend). This translates to near real-time inference latency (a few seconds) when deployed on modern data-center GPUs, satisfying the rigorous constraints of operational ice charting.

\subsection{Implementation and Training Details}
The base U-Net model was implemented using the PyTorch deep learning framework. During the training phase, we utilized the Adam optimizer with an initial learning rate of $10^{-4}$, dynamically reduced via a learning rate scheduler to ensure smooth convergence. Due to the massive spatial extent of the original SAR scenes, the training data was dynamically sampled into $256 \times 256$ pixel patches. 

To optimize training efficiency and accommodate the irregular shapes of the WMO Egg Code polygons, the strict regional constraint was relaxed to a patch-level formulation. During training, $256 \times 256$ pixel patches are dynamically extracted. The target macroscopic concentration $C_p$ for each patch is computed as the area-weighted average of the overlapping polygons. Consequently, the network applies a spatial Global Average Pooling (GAP) over the entire patch to predict this aggregated regional concentration. This architectural trade-off ensures rapid convergence and avoids the severe memory bottlenecks associated with dynamic boolean masking, all while preserving the underlying macroscopic statistical fidelity. Furthermore, to prevent the network from overfitting to the low-resolution structural biases of the weak labels, data augmentation techniques---including random rotations and flips---were applied.

\subsection{Ablation Study: Robustness of Analytical Scaling}

\subsubsection{Impact of Spatial Regularization}
SAR imagery is notoriously affected by multiplicative speckle noise, which propagates through the network's convolutions and introduces high-frequency artifacts in the latent logit space $Z$. In an ablation experiment where the spatial low-pass filter (Gaussian Blur) was omitted prior to scaling, the dynamic percentile calculation was disproportionately influenced by these speckle artifacts. This resulted in a ``salt-and-pepper'' binarization where pure open water regions contained erroneous micro-floes. The spatial regularization step is therefore mandatory to maintain topological consistency before applying the rigid saturation bounds.

\subsubsection{Sensitivity to Percentile Selection}
The choice of the 2nd and 98th percentiles ($z_{2\%}$ and $z_{98\%}$) instead of the absolute extrema is a critical design choice for robust calibration. Using the absolute minimum and maximum (0th and 100th percentiles) makes the analytical scaling highly vulnerable to point-target anomalies. Specifically, elements such as ships, static icebergs, or extreme sensor noise create massive local logit spikes. These anomalies artificially stretch the dynamic range, heavily compressing the logits of the actual sea ice and neutralizing the binarization effect. 

Conversely, using tighter bounds (e.g., 10th and 90th percentiles) prematurely saturates the sigmoid function, destroying the fine-grained probability gradients at the edges of the leads. Empirical observations confirm that trimming the top and bottom 2\% of the latent distribution provides the optimal statistical trade-off between outlier rejection and contrast maximization.

\section{Experiments and Results}

\subsection{Qualitative Evaluation: Revealing Hidden Topology}
\begin{figure*}[!t]
\centering
\includegraphics[width=0.93\textwidth]{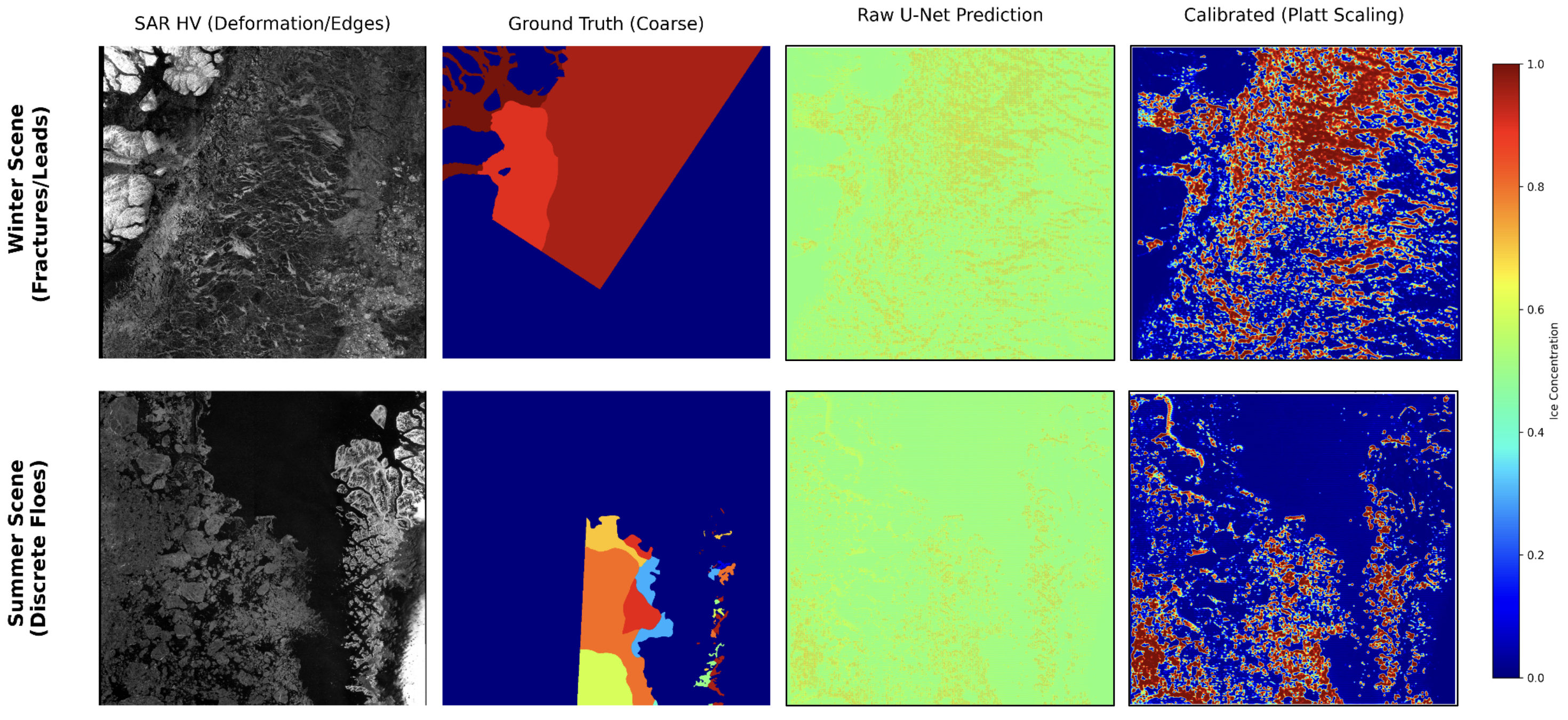}
\caption{Comparison of segmentation results. (Col 1) SAR HV input showing physical topology. (Col 2) Regional weak labels from Ice Charts. (Col 3) Uncalibrated U-Net predictions exhibiting severe under-confidence. (Col 4) Our Analytical Logit Scaling revealing high-resolution leads and discrete ice floes without manual pixel-level supervision. Note: The shared colorbar represents macroscopic Sea Ice Concentration for the Ground Truth. For the calibrated prediction, the Platt Scaling pushes values to the sigmoid's saturation bounds, effectively acting as a binary topological mask.}
\vspace{-0.4cm} % Remonte le texte de 4 millimètres
\label{fig:results}
\end{figure*}

To evaluate the topological extraction capability of our method, we first analyze a winter scene characterized by compact pack ice intersecting with narrow leads, as illustrated in Fig. \ref{fig:results}. Under standard inference, the weakly supervised U-Net minimizes the regional loss by outputting a uniform, blurry probability field, completely masking the fractures (see Fig. \ref{fig:results}, Col 3).

By applying the Analytical Logit Scaling, the latent representations are stretched to the sigmoid's saturation bounds. As shown in the final column of Fig. \ref{fig:results}, this dynamic scaling snaps pixels to 0 or 1 based on their relative physical backscatter, suddenly revealing the intricate network of leads that were physically present in the SAR imagery but mathematically suppressed by the regional loss constraint.

\subsection{Quantitative Evaluation: The Paradox of the Mean}
The robustness of the proposed scaling is further demonstrated on a highly fragmented summer scene within the marginal ice zone, as detailed in Table \ref{tab:metrics}. For instance, focusing on Polygon 22, the ground truth Ice Chart assigns a macroscopic concentration of exactly 30\% ($C_p = 0.30$) to the region. The uncalibrated network conservatively predicts values around $0.30$ for almost every pixel to minimize the loss. 

After applying our analytical scaling, the network outputs discrete binary ice floes and open water. Remarkably, when these binarized pixels are averaged over the polygon, the reconstructed macroscopic concentration is $0.296$ (Table \ref{tab:metrics}). This perfectly satisfies the original regional constraint while providing high-resolution topological boundaries. When evaluating the resulting binary map against the complex fragmented ice distribution, the model achieves an overall accuracy of 77.7\%. This proves that the method successfully resolves local topology without sacrificing regional statistical fidelity.

\begin{table}[!t]
\renewcommand{\arraystretch}{1.3}
\caption{Consistency between ground-truth regional concentration ($C_p$) and our aggregated binary predictions on selected polygons from the summer scene. The overall scene accuracy reaches 77.7\%.}
\label{tab:metrics}
\centering
\begin{tabular}{lccc}
\hline
\textbf{Polygon ID} & \textbf{Target ($C_p$)} & \textbf{Predicted} & \textbf{Abs. Error} \\
\hline
22 & 0.30 & 0.296 & 0.004 \\
31 & 0.30 & 0.349 & 0.049 \\
43 & 0.60 & 0.527 & 0.073 \\
44 & 0.80 & 0.695 & 0.105 \\
\hline
\end{tabular}
\vspace{-0.4cm}
\end{table}

\section{Discussion and Limitations}
While highly effective for topological extraction, our approach fundamentally alters the probabilistic nature of the network. Standard Temperature Scaling is designed to calibrate model confidence, ensuring that a predicted probability of 0.8 reflects an 80\% likelihood of correctness. Conversely, our analytical scaling explicitly destroys this uncertainty estimation by forcing extreme overconfidence (mapping logits to $[-5, 5]$). Consequently, the output should be interpreted as a deterministic topological feature extractor---an adaptive spatial contrast enhancer---rather than a true probabilistic confidence map.

\section{Conclusion}
In this paper, we addressed the fundamental scale mismatch in weakly supervised sea ice segmentation caused by regional WMO Egg Code labels. We introduced a tuning-free, post-inference Analytical Logit Scaling method that globally stretches the raw output logits based on scene-specific percentiles. This approach successfully forces physical binarization, extracting high-resolution (40m) topology such as leads and discrete ice floes without requiring any expensive, manual pixel-level annotations.

However, our analysis also highlights a persistent challenge inherent to weakly supervised approach: for some patches, our current pipeline occasionally focuses on leads to predict the concentration of sea-ice. This ambiguity could be lifted by training the model on all the images of the AI4Arctic challenge. Despite this limitation, by preserving regional concentration metrics while revealing hidden local structures, this lightweight pipeline successfully bridges the gap between scalable weak supervision and high-fidelity operational sea ice charting, paving the way for future multimodal approaches to resolve these remaining physical ambiguities.

% \enlargethispage{1cm}

\bibliographystyle{IEEEtran}
\bibliography{refs}

\end{document}